\documentclass{llncs}
\usepackage{caption}
\usepackage{graphicx} 
\usepackage{hyperref}
\usepackage{multirow}
\usepackage{subcaption}
\usepackage{tikz}
\usetikzlibrary{shapes, arrows, positioning}
\usepackage[capitalise,nameinlink]{cleveref}
\usepackage[ruled,vlined]{algorithm2e}
\begin{document}

\title{PELP: Pioneer Event Log Prediction Using Sequence-to-Sequence Neural Networks}
\author{Wenjun Zhou\orcidID{0000-0002-2262-1428} \and
Artem Polyvyanyy\orcidID{0000-0002-7672-1643} \and
James Bailey}
\authorrunning{W. Zhou et al.}
%
\institute{The University of Melbourne, Victoria VIC 3010, Australia\\
 \email{\{wenjun.zhou,artem.polyvyanyy,baileyj\}@unimelb.edu.au}}
\maketitle              
\begin{abstract}
Process mining, a data-driven approach for analyzing, visualizing, and improving business processes using event logs, has emerged as a powerful technique in the field of business process management. 
Process forecasting is a sub-field of process mining that studies how to predict future processes and process models.
In this paper, we introduce and motivate the problem of \emph{event log prediction}, and present our approach to solving the event log prediction problem, in particular, using the sequence-to-sequence deep learning approach. 
We evaluate and analyze the prediction outcomes on a variety of synthetic logs and seven real-life logs and show that our approach can generate perfect predictions on synthetic logs and that deep learning techniques have the potential to be applied in real-world event log prediction tasks. 
We further provide practical recommendations for event log predictions grounded in the outcomes of the conducted experiments.
\keywords{Event log prediction \and process mining \and deep learning \and Seq2Seq.}
\end{abstract}

\section{Introduction}
\label{Introduction}
Process mining is an emerging interdisciplinary research field that resides at the intersection of data science and process science~\cite{DBLP:books/sp/Aalst16}. 
Through process mining, analysts can identify process bottlenecks and noncompliance during the process execution to improve the process using its visual representations, such as Directly-Follows Graphs (DFGs), Petri Nets, and Process Trees~\cite{DBLP:books/sp/Aalst16}. 
These models are abstractions of the processes they represent, where event logs collected through a software system are used as the input for generating the models. 
This approach of transforming event logs into process models is also known as process discovery, a sub-field of process mining.

With the business demand for prediction of the future and the prospering of machine learning in recent years, researchers have started a trend of predicting process elements. 
For example, Cardoso, J. and Leni\v{c}, M.~\cite{firstActivityPrediction} proposed an approach to business activity prediction. 
A line of research has focused on predicting time aspects in processes~\cite{RN140}. 
Existing techniques, however, focus on predicting case-level process elements, with the prerequisite that a process case has executed a few activities~\cite{RN656,RN2863}. 
Very little research has focused on predicting future event logs or models. 
With the concept being proposed by Poll et. al.~\cite{DBLP:conf/bpm/PollPRRR18}, one technique has been devised to forecast future process models~\cite{DBLP:conf/er/SmedtYPWM21,Smedt2023}.

Even though businesses can use simulations to generate de facto future logs from the discovered process models, the challenge is that the discovered process model is a representation of the past with a certain level of generalization to the future, but it may have less focus on the dynamics of the model itself. 
In other words, the discovered models have a weaker ability to anticipate the true future other than simulations. 
This is due to the misinterpretation of the generalization of models~\cite{DBLP:journals/topnoc/SyringTA19}. 
If a future process model or an event log that describes the future as closely as possible can be predicted, this will contribute to companies and organizations in terms of foreseeing upcoming changes and, hence, support better business planning.

This paper presents the Pioneer Event Log Prediction (PELP) approach grounded in deep learning, a Sequence-to-Sequence (Seq2Seq) neural network architecture in particular. 
The training input and output pairs are prepared in occurrence order to allow the neural network to learn the dependencies between the input and its immediate future output. 
Eventually, after training, the trained neural network takes the most recent historical event logs as input and produces the future event logs, that is, the event log the system that executes the process will generate in the future. 
In particular, we contribute the following:
\begin{itemize}
\item A framework for predicting the future event log (the PELP framework) based on a given event log of historical process executions grounded in deep neural networks.
\item A realization of the framework using Seq2Seq neural networks.
\item An evaluation of the proposed event log prediction technique over synthetic and real-world event logs.
\item A discussion of event log characteristics that suggest the use of particular prediction methods.
\end{itemize}

The following section specifies the preliminary terms and background knowledge for readers to understand the remaining content, while \cref{Related Work} discusses related work.
\Cref{Log Prediction} presents the framework for event log prediction using deep learning neural networks, and its subsections describe the detailed approach taken to predict future event logs using Seq2Seq architecture, followed by \cref{eval}, providing evaluations and analysis of the proposed approach. 
Finally, \cref{Conclusion} concludes the paper.

\section{Preliminaries}
\label{Preliminaries}
\paragraph{\textbf{Seq2Seq.}}
Sequence-to-Sequence (Seq2Seq)~\cite{DBLP:conf/nips/SutskeverVL14} is a deep learning neural network architecture capable of processing a sequence of input tokens and producing a sequence of output tokens. 
It is widely used in many applications, such as machine translation and time series predictions~\cite{che2016recurrent,10.5555/2969033.2969173}. 
This architecture comprises an encoder and a decoder. 
During training, the encoder transforms input sequences into context vectors, and the decoder uses the information in the context vectors to determine the output tokens with respect to the input sequence. 
Typical neural networks, such as Recurrent Neural Network (RNN), Long Short-Term Memory (LSTM), or Gated Recurrent Unit (GRU), can be used as the encoder and decoder. 
In the experiments conducted in this article, a single GRU layer encoder and a single GRU layer decoder with attention were used. 
Attention~\cite{bahdanau2016neural} is a mechanism that makes the model focus on the ``important'' parts of the input data. 
We use attention techniques in our decoder to filter out possible noises in the input sequences. 
DLM specifically refers to the deep learning model for the model weights and parameters learned during training.

\paragraph{\textbf{Event logs.}}
In process mining, an \emph{event log}, or a \emph{log}, captures the activities taken during a business process. 
Each log consists of multiple events. 
At least three compulsory attributes are recorded for each event: \emph{activity}, \emph{case identifier} (case ID), and \emph{timestamp}. 
The \emph{activity} attribute refers to the executed activity instance. 
The \emph{timestamp} records the time of the corresponding activity occurrence. 
Finally, all activities from the same business process instance share the same \emph{case ID} attribute. 
There are two ways in which the logs are referred to in this paper. 
A \emph{raw log} is expressed as a set of data triplets, where the first field in the triplet represents the \emph{case ID}, the second field represents the \emph{timestamp}, followed by the \emph{activity} in the last field. 
For example, $\{(2,6,d),$ $(2,5,b),$ $(1,1,a),$ $(1,7,c),$ $(2,2,a),$ $(2,4,b),$ $(1,3,b)\}$ is an example of a raw log $L$ of seven events.

Using these compulsory event attributes, one can construct the directly-follows relation of the \emph{log}. 
\emph{Activities} that refer to the same \emph{case ID} ordered by their \emph{timestamps} from the earliest to the latest form a \emph{trace}. 
An \emph{event log} can keep records on multiple \emph{traces}. 
The previous example log $L$ specifies two traces and, thus, can be given as the unordered set $\{\langle a, b, c\rangle$, $\langle a, b, b, d\rangle\}$. 
If the same trace appears multiple times in a log, one can use a multi-set to capture it, for instance $\{\langle a, b, c\rangle$, $\langle a, b, b, d\rangle^2\}$, where the superscripts denotes the numbers of occurrences of that trace variants; note that in this log, differently from log $L$, trace $\langle a, b, b, d\rangle\}$ occurs twice. 
To express the ordered \emph{traces}, we use parentheses instead of curly braces, while for reoccurring patterns, we use square brackets to enclose the repeating part followed by an ellipsis to indicate that the pattern repeats infinitely. 
To illustrate, $([\langle a, b, c\rangle$, $\langle a, b, b, d\rangle^2], \ldots)$ represents an ordered trace list which starts with a trace $\langle a, b, c\rangle$ (the superscript 1 is omitted in this case), followed by two $\langle a, b, b, d\rangle$ traces, and this pattern repeats infinitely.
Traces form an event log can be ordered using different rules. 
In our work, in particular, we order the traces based on the occurrence of the first activity in each trace.

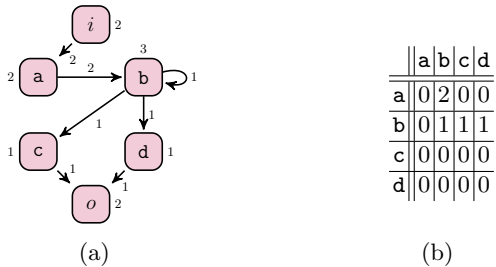
\begin{figure}[ht]
\centering
\subcaptionbox{\scriptsize \label{fig:dfgR}}{
\begin{tikzpicture}[scale=0.55, transform shape, ->, >=stealth', shorten >=1pt, auto, node distance=18mm, on grid, semithick, action/.style={fill=purple!20, draw, rounded corners, minimum size=9mm}]
\node[action,label=0:\small 2]    (n0) 						{\Large $i$};
\node[action,label=180:\small 2]  (n1) [below left=of n0]   {\Large $\texttt{a}$};
\node[action,label=90:\small 3]   (n2) [below right=of n0]  {\Large $\texttt{b}$};
\node[action,label=180:\small 1]  (n3) [below=of n1]        {\Large $\texttt{c}$};
\node[action,label=0:\small 1]    (n4) [below=of n2]		{\Large $\texttt{d}$};
\node[action,label=0:\small 2]    (n5) [below right=of n3]  {\Large $o$};
\path (n0) edge node {\small 2} (n1)
		(n1) edge node {\small 2} (n2)
		(n2) edge node {\small 1} (n3)
				 edge node {\small 1} (n4)
				 edge [loop right] node {\small 1} (n3)
		(n3) edge node {\small 1} (n5)
		(n4) edge node {\small 1} (n5)
;
\end{tikzpicture}
}
\hspace{60pt}
\subcaptionbox{\normalsize \label{fig:dfmR}}{
\begin{tabular}[b]{c||c|c|c|c}
& \texttt{a} & \texttt{b} & \texttt{c} & \texttt{d} \\ \hline\hline
\texttt{a} & 0 & 2 & 0 & 0 \\ \hline
\texttt{b} & 0 & 1 & 1 & 1 \\ \hline
\texttt{c} & 0 & 0 & 0 & 0 \\ \hline
\texttt{d} & 0 & 0 & 0 & 0
\end{tabular}
\vspace{3mm}
}
\caption{(a) DFG and (b) directly-follows matrix of relation $R$.}
\label{fig:dfgR&dfmR}
\end{figure}

\paragraph{\textbf{Directly-Follows Relationships.}}
The \emph{directly-follows relation} of an event log is a set of pairs of activities. 
The activity pair comes in consecutive order, in which the first activity happened at time \emph{t}, and the second happened at a later time \emph{t+n}, where \emph{n} can be any time interval greater than zero.
A pair is in the relation only if the log specifies a trace in which there are two consecutive events such that the preceding event specifies an occurrence of the first activity in the pair and the subsequent event captures an occurrence of the second activity.
Log $L$ specifies two traces $\langle a, b, c\rangle$ and $\langle a, b, b, d\rangle$ and the directly-follows relation $R =$ $\{(a,b) ,(b,b) ,(b,c) ,(b,d)\}$.
One can construct a Directly-Follows Graph (DFG)~\cite{DBLP:books/sp/Aalst16} or a directly-follows matrix from a given directly-follows relation. 
In this paper, we use weighted adjacency matrices to measure the quality of the prediction and quantify the distance from our prediction and ground truth.
\Cref{fig:dfgR,fig:dfmR} show the DFG and the directly-follows matrix of relation $R$.
\vspace{-8pt}
\vspace{-19pt}
\paragraph{\textbf{RMSE and MAE.}}
Root Mean Square Error (RMSE), sometimes also referred to as Root Mean Square Deviation, and Mean Absolute Error (MAE) are measures of the differences between values~\cite{Hodson:2022aa}.
We calculate RMSE and MAE between two square matrices $X$ and $Y$ of size $n$ as shown below.
RMSE and MAE are often used to measure the differences between two DFGs or event logs~\cite{DBLP:conf/er/SmedtYPWM21,Smedt2023}.

\begin{equation}\label{rmse&mae}
    \mathit{RMSE} = \sqrt{\frac{1}{n^2}\sum_{i=1}^{n}\sum_{j=1}^{n}\left(X_{ij}-Y_{ij}\right)^2}
    \quad \quad \mathit{MAE} = \frac{1}{n^2}\sum_{i=1}^{n}\sum_{j=1}^{n}\left|X_{ij}-Y_{ij}\right|
\end{equation}

\section{Related Work}
\label{Related Work}
In this section, we summarise publications from the domain of predictive process improvement.
We also compare the differences between our work and the existing publications. 
Two sub-areas study predictions of process artifacts, namely Predictive Process Monitoring (PPM)~\cite{RN656} and Process Forecasting~\cite{DBLP:conf/bpm/PollPRRR18}. 
PPM focus is on case-, or micro-level, predictions. 
In PPM, outcome prediction, next activity prediction, and process duration prediction are the main use cases that operate within the scope of a single process case. 
In contrast, Process Forecasting, including process model prediction and event log prediction, focuses on the model-, or macro-level, predictions. 

In general, the vast majority of PPM prediction techniques take historical event logs and learn a model encoding possible process case behaviors. 
Then, the newly lodged process case is monitored, and the so far observed prefix of the case is used to generate predictions for that particular case. 
The benefit of PPM is that the prediction for a particular case can often be made in real-time, and the newly observed data can be used immediately to update the learned models to improve future predictions.
PPM techniques are often deployed before they get used to allow sufficient training before the techniques get productive.
The state-of-the-art PPM techniques often demonstrate high prediction accuracies~\cite{RN140,RN2484,RN1584} and are grounded in conventional statistical and process analysis techniques, while several existing works also explore deep learning approaches~\cite{RN1058,RN50,RN565,RN1572}.

Process Forecasting studies changes in process models over time. The concept was proposed in 2018; since then, only a few techniques have been proposed. 
The work by De Smedt et. al.~\cite{DBLP:conf/er/SmedtYPWM21,Smedt2023}, explores how statistical methods over time series help in predicting the directly-follows relationships of real-life business processes. 
The lesson learned from this work is that no method works well across all datasets. 
However, the quality of achieved predictions can enable proactive business process planning, including process drift and change predictions. 

Our work contributes to the Process Forecasting line of research, as its predictions focus on macro-level process model behavior. 
For example, compared to the deep learning approaches in PPM, the granularity of the training data in our approach is different.
We use a sequence of ordered cases as training data, while PPM techniques use individual cases for training.
The difference between our work and the one by De Smedt et al. is that their approach takes a directly-follows matrix as input for training and outputs the future directly-follows matrix as output, while our technique takes a pre-processed event log as input and outputs the predicted event log, thus aiming to achieve more fine-grained predictions.

\section{Log Prediction}
\label{Log Prediction}
\begin{figure}[t]
\vspace{8pt}
    \includegraphics[width=\textwidth]{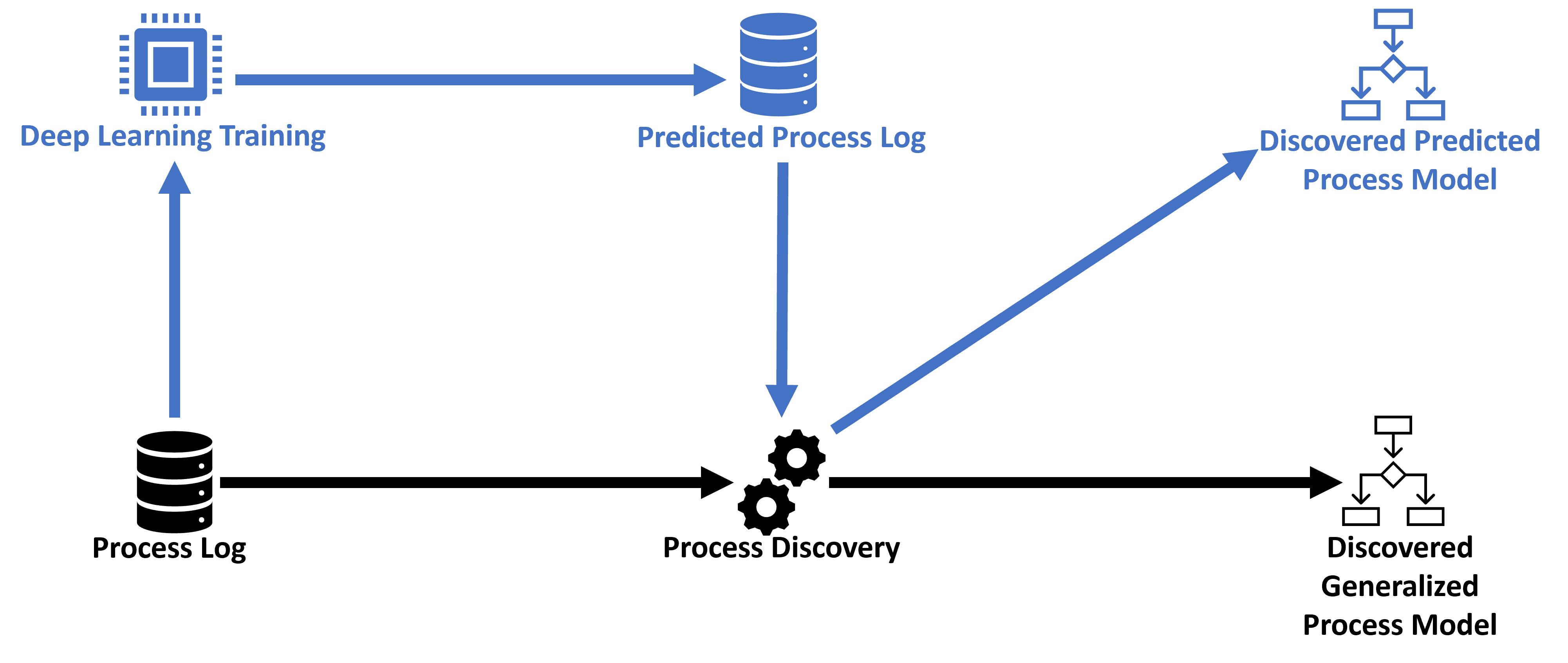}
    \caption{Relationships between event log prediction and process discovery.} \label{fig:process-log-prediction-comparison}
\vspace{-8pt}
\end{figure}

Next, we propose the PELP framework for event log prediction, which complements PPM techniques~\cite{RN656,RN2863}. 
The event log prediction we propose is a black-box approach combining process discovery and process simulation principles. 
\Cref{fig:process-log-prediction-comparison} describes the relationship between event log prediction and conventional process discovery.
The conventional process discovery techniques aim to construct a so-called Generalized Process Model (GPM), which is a model that describes both the historical and future process cases well. 
This implies that the discovery algorithm is supposed to generalize well enough to serve the two purposes. 
Indeed, if the underlying process does not change over time, a good generalized model will serve for prediction purposes well.
However, we argue that the GPM may struggle to be good enough and may not reflect the true process of a system for a specified time frame if the business model changes over time.

\begin{figure}[h]
\vspace{-8pt}
    \includegraphics[width=\textwidth]{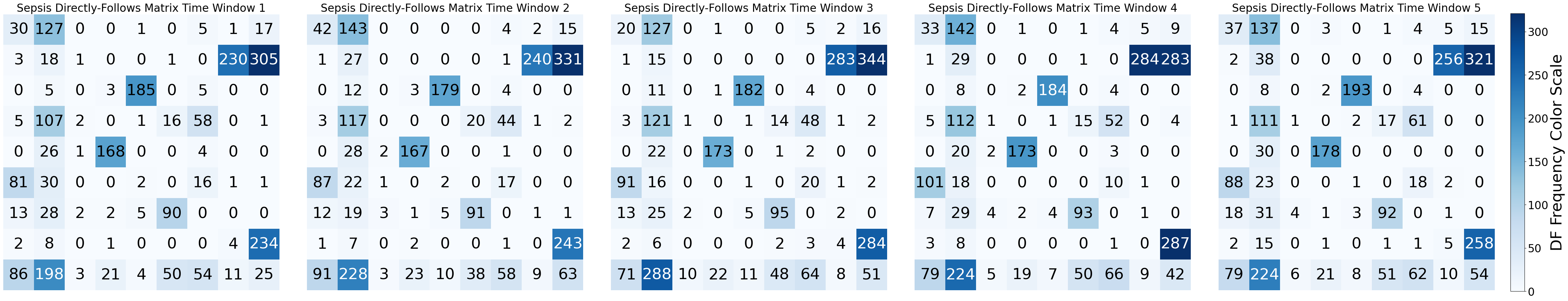}
    \caption{Process model frequency change over time (Sepsis).} \label{fig:process-model-frequency-change-over-time}
\vspace{-24pt}
\end{figure}

\begin{figure}[h]
\vspace{-8pt}
    \includegraphics[width=\textwidth]{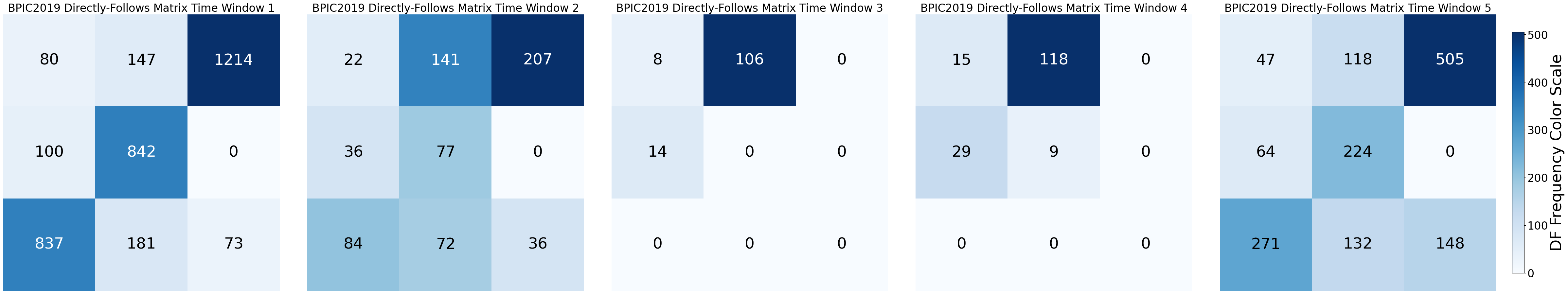}
    \caption{Process model structure change over time (BPIC2019).} \label{fig:process-model-structure-change-over-time}
\vspace{-12pt}
\end{figure}

To illustrate, we take part in the real-life event log, Sepsis Cases Event Log\footnote{\url{https://doi.org/10.4121/uuid:915d2bfb-7e84-49ad-a286-dc35f063a460}}, as an example, sort the traces over time based on the occurrences of the first activity in each trace, select the first one thousand traces and split them into five equal amounts. 
The directly-follows relationship heat map of an interesting area in the dataset is shown in \cref{fig:process-model-frequency-change-over-time}.
As shown in the figure, structure-wise, there is no change over time; by structure, we mean that more than 50\% of the observed directly-follows relationships do not display significant frequency change over the chosen time window. 
While frequency-wise, we observed changes in some directly-follows relationships over time; by frequency, we mean the observed directly-follows relationships display a minor to moderate level of change within about 20\% difference from the previous time window. 
On the other hand, \cref{fig:process-model-structure-change-over-time} shows the structure change over time in the interesting area of the partial BPIC2019 event log.\footnote{\url{https://doi.org/10.4121/uuid:d06aff4b-79f0-45e6-8ec8-e19730c248f1}}

\begin{figure}[t]
    \includegraphics[width=\textwidth]{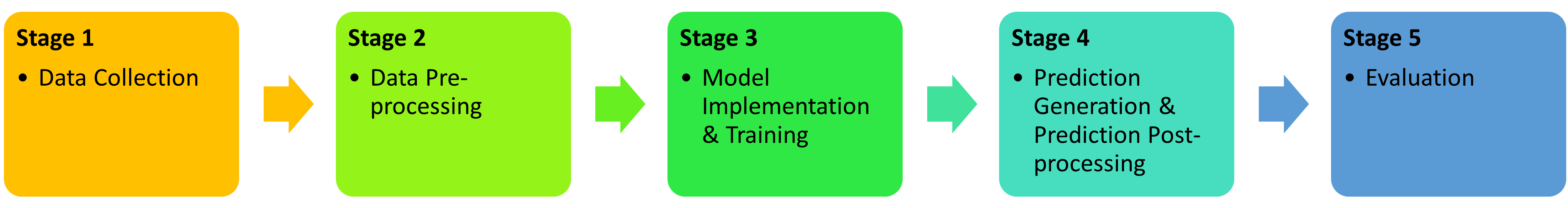}
    \caption{Five-stage of the PELP framework for event log prediction.} \label{fig:5stage-log-prediction}
    \vspace{-18pt}
\end{figure}

The PELP framework for event log prediction comprises five stages, namely data collection, data pre-processing, model implementation and training, prediction generation and post-processing, and evaluations; refer to \cref{fig:5stage-log-prediction}. 
For each stage, there are several choices that may impact the prediction outcome and performance. 
We describe each stage and our approach to each stage in the following subsections.

\subsection{Data Collection}
\label{stage1}
Event logs collected should comply with XES standard~\cite{10267858}, which requires every event to have at least three mandatory fields: case ID, activity, and timestamp. Using other optional payload fields, based on the analysts' domain knowledge at design time, one may consider collecting other process-relevant information through the process software systems or sensors. If the data collection can be controlled, one can consider using different data formats for recording the data values. Data collected in a format suitable for the target prediction model can reduce the effort of data pre-processing. For our experiments, we selected seven suitable event logs with various characteristics from the Task Force on Process Mining (TFPM) website\footnote{\url{ https://www.tf-pm.org/resources/logs}}. These logs are selected due to their public availability, standard format (xes or csv), a sufficient number of traces for DLM training tasks, and availability of the mandatory fields. The logs we selected are: 
BPI Challenge 2011---Hospital Log (BPIC2011)\footnote{\url{https://data.4tu.nl/articles/_/12716513/1}},
BPI Challenge 2012 (BPIC2012)\footnote{\url{https://doi.org/10.4121/uuid:3926db30-f712-4394-aebc-75976070e91f}}, 
BPI Challenge 2017 (BPIC2017)\footnote{\url{https://doi.org/10.4121/14625948.v1}},
BPI Challenge 2019 (BPIC2019), 
Sepsis Cases---Event Log (Sepsis),
Hospital Billing Event Log (HB)\footnote{\url{https://doi.org/10.4121/uuid:76c46b83-c930-4798-a1c9-4be94dfeb741}}, and 
Road Traffic Fine Management Process (RTFMP)\footnote{\url{https://doi.org/10.4121/uuid:270fd440-1057-4fb9-89a9-b699b47990f5}}. 
The characteristics of these selected event logs are summarized in \cref{pmtf_log}.

\subsection{Data Pre-processing}
\label{stage2}
Once the data is collected, it should be pre-processed, that is, the data should be converted to a format that is readable and processable. 
Any incomplete data should be removed or filled with synthetic data. 
Depending on the design and the prediction objectives, data at this stage is processed in an $(X,Y)$ pair format, where $X$ is used as training input to feed the model, and $Y$ is used as the expected output to guide the update of the gradients of the DLM.

Because not all the real-life logs contain optional fields and the number of optional fields and their data types in different logs are different, which complicates the batch pre-processing of all logs, we decided to only use the three mandatory fields for training. 
Given an event log from the previous stage, in this stage, we order events for each case ID based on their timestamps; if two events have the same timestamp, we sort them based on the activities they refer to (in alphabetical order). 
We then order the traces based on the timestamps of their first events. 
If two traces have first events with the same timestamp, we order them based on the alphabetical order of their activities. 
We remove special characters and symbols (such as whitespaces and hyphens) from the activities before sorting. 
Next, we write the pre-processed log into a text file, in which the number of lines matches the number of traces in the log, and each line specifies the sequence of activities of that trace separated by whitespace characters.

\begin{table}[t]
\vspace{-16pt}
    \centering
    \caption{Event log characteristics (selected in the experiments/full event log); ``CL'' is the abbreviation of ``case length''.}\label{pmtf_log}
    \begin{tabular}{|l|r|r|r|r|}
    \hline
    {\bfseries Log} & {\bfseries Total cases\,\,\,} & {\bfseries Total activities\,\,\,} & {\bfseries Unique cases\,\,\,} & {\bfseries Unique activities}\\
    \hline
    BPIC2011 & 1,000/1,143\,\,\, & 133,586/150,291\,\,\, & 847/982\,\,\, & 603/623 \\
    BPIC2012 & 1,000/13,087\,\,\, & 21,902/262,200\,\,\, & 470/3,792\,\,\, & 24/24 \\
    BPIC2017 & 1,000/31,509\,\,\, & 39,932/1,202,267\,\,\, & 800/15,930\,\,\, & 24/26 \\
    BPIC2019 & 1,000/251,734\,\,\, & 14,832/1,595,923\,\,\, & 213/11,825\,\,\, & 33/42 \\
    HB & 1,000/100,000\,\,\, & 5,042/451,359\,\,\, & 62/1,027\,\,\, & 15/18 \\
    Sepsis & 1,000/1,050\,\,\, & 14,481/15,214\,\,\, & 658/691\,\,\, & 16/16 \\
    RTFMP & 1,000/150,370\,\,\, & 4,501/561,470\,\,\, & 25/260\,\,\, & 11/11 \\
    \hline
    {\bfseries Log} & {\bfseries Maximum CL\,\,\,} & {\bfseries Minimum CL\,\,\,} & {\bfseries Average CL\,\,\,} & {\bfseries Median CL}\\
    \hline
    BPIC2011 & 1,814/1,814\,\,\, & 1/1\,\,\, & 133.59/131.49\,\,\, & 50/55 \\
    BPIC2012 & 130/175\,\,\, & 3/3\,\,\, & 21.90/20.035\,\,\, & 16/11 \\
    BPIC2017 & 124/180\,\,\, & 12/10\,\,\, & 39.93/38.156\,\,\, & 36/35 \\
    BPIC2019 & 990/990\,\,\, & 2/1\,\,\, & 14.83/6.34\,\,\, & 6/5 \\
    HB & 25/217\,\,\, & 1/1\,\,\, & 5.04/4.51\,\,\, & 6/5 \\
    Sepsis & 185/185\,\,\, & 3/3\,\,\, & 14.48/14.49\,\,\, & 13/13 \\
    RTFMP & 9/20\,\,\, & 2/2\,\,\, & 4.50/3.73\,\,\, & 5/5 \\
    \hline
    \end{tabular}
    \vspace{-24pt}
\end{table}

We then prepare the training pairs $(X,Y)$. 
The prediction problem is abstracted to a transformation that, given historical log traces, predicts future traces. 
Hence, we parameterize the number of traces concatenated in training input $X$ and expected prediction output $Y$. 
To distinguish traces within the input and output sequences, we append an end of trace ``$\mathit{EOT}$" token at the end of each trace during trace concatenation. 
For example, consider the experiment setup that uses three traces to predict two traces in the future and the ordered input file shown in \cref{logtxt}; to improve readability, we use commas (,) between the activities instead of whitespaces. 
If $(X_1,Y_1)$ represents the first training pair, then $X_1$ is equal to $a1$, $a2$, $\mathit{EOT}$, $b1$, $\mathit{EOT}$, $c1$, $c2$, $c3$, $\mathit{EOT}$ and $Y_1$ is equal to $d1$, $d2$, $\mathit{EOT}$, $e1$, $e2$, $\mathit{EOT}$. 
For the next training pair $(X_2,Y_2)$, $ X_2$ is equal to $b1$, $\mathit{EOT}$, $c1$, $c2$, $c3$, $\mathit{EOT}$, $d1$, $d2$, $\mathit{EOT}$, and $Y_2$ is equal to $e1$, $e2$, $\mathit{EOT}$, $f1$, $f2$, $f3$, $\mathit{EOT}$, respectively. 
This example event log does not define further training pairs, as it does not contain further subsequences of five consecutive traces.

\begin{table}[t]
    \vspace{-8pt}
    \centering
    \caption{Example pre-processed event log.}\label{logtxt}
    \begin{tabular}{|l|}
    \hline
    $a1, a2$ \\
    $b1$ \\
    $c1, c2, c3$ \\
    $d1, d2$ \\
    $e1, e2$ \\
    $f1, f2, f3$ \\
    \hline
    \end{tabular}
    \vspace{-20pt}
\end{table}

The Seq2Seq DLM is a suitable selection for our prediction task as the lengths of the training pairs vary, and the Seq2Seq DLM can handle this variation. 
It is worth noting that one may choose a different pre-processing approach other than ours. 
We have selected the approach that is the most intuitive, ordering the cases based on the occurrence of their first activity; while one can order the cases based on the earliest last activity, or the middle point of the case duration, or even only ordering the activities based on their timestamps and reconstruct case information (Case ID) later in the post-processing stage.

\subsection{Model Implementation and Training}
\label{stage3}
As indicated in the previous section, the DLM selected for the prediction task should be based on the suitability of the prediction objectives, e.g., classification, regression, or goal recognition, and the nature of the data properties. 
The DLM may include several hyper-parameters depending on the design and implementation. 
The DLM we use in this work has a fixed architecture consisting of an encoder and a decoder with attention, both having an embedding layer and a single GRU layer. 
As we do not know which hyper-parameter combinations are the most effective for the data set, we implemented two hyper-parameter search methods: grid search and random search. 
To reduce the search space, we fixed some hyper-parameters based on our experience\footnote{This experience stems from the experiments with the synthetic datasets; refer to \cref{eval} for details.}: always use Stochastic Gradient Descent (SGD) as optimizer; always use cross-entropy as the loss function; always use teacher forcing; the model is always initialized with zeros; vary the learning rate from 0.001 to 0.3; vary the hidden layer size from 16 to 1024; vary the dropout probability from 0.001 to 0.3. For more detailed descriptions of these hyper-parameters, one can refer to \emph{``Hyper-Parameter Optimization: A Review of Algorithms and Applications"}~\cite{yu2020hyperparameter}.
Finally, we customized the number of traces concatenated in training pairs; the input and the output vary from 1 to 50 concatenated traces. 
Our DLM takes the input sequence, splits it by activities, and outputs one token at a time until it reaches an internally defined stop signal. 
The ``$\mathit{EOT}$" token is treated the same as an activity in the model. 
The model is trained until the loss has not been improved for one hundred epochs.

Due to the low training efficiency of some DLMs, such as Seq2Seq, especially for event logs with long median or average trace length, such as BPIC2011 and BPIC2017, and for event logs with an abundant number of training records, like RTFMP, BPIC2019 and BPIC2012, in the experiments, for each event log, we only select the first one thousand traces after ordering all the traces in the entire log based on the timestamps of traces' first events.
To test our prediction approach, we split the input event log into two parts: the first is used for training the DLM and the second is used as the testing data. 
We use the first 80 percent of the traces in the input event log (supplied in the text format) for training and the remaining 20 percent for testing.

\subsection{Prediction Generation and Post-processing}
\label{stage4}
After training, and depending on the model saving strategy, the saved selected model is reloaded for the prediction task. Once the prediction is obtained for evaluation, an appropriate evaluation method should be selected. Here, two aspects need to be specifically pointed out:
1) The quality measures used to evaluate the performance of predictions should be fair to the corresponding trained DLM. For example, if cross-entropy loss is used as the loss function to train the DLM, the quality measure should aim to quantify aspects related to those penalized by cross-entropy loss. 
2) After obtaining predictions, one can directly evaluate the results based on the selected quality measure or post-process the predictions to suit different measurement purposes. 
In other words, post-processing is optional depending on one's measurement.

When training is finished, we use the last \textit{N} training traces as the input for the model to generate its predicted output traces. 
Consider the example event log from \cref{stage2}. 
To generate the first predicted traces, we concatenate the last three traces in the log as the input sequence to the trained Seq2Seq model.
Take \cref{logtxt} as an example, recall that we used three traces to predict two traces; to generate the following predicted traces after training, we would use $\mathit{EOT}$, $d1$, $d2$, $\mathit{EOT}$, $e1$, $e2$, $\mathit{EOT}$, $f1$, $f2$, $f3$, $\mathit{EOT}$ to generate prediction $g1$, $g2$, $g3$, $\mathit{EOT}$, $h1$, $h2$, $\mathit{EOT}$. Note that $g1$, $g2$, $g3$, $\mathit{EOT}$, $h1$, $h2$, $\mathit{EOT}$ here are imaginary, the actual output will depend on the model's learning. If we wish to continue the prediction, we would use $\mathit{EOT}$, $f1$, $f2$, $f3$, $\mathit{EOT}$, $g1$, $g2$, $g3$, $\mathit{EOT}$, $h1$, $h2$, $\mathit{EOT}$ to generate the next two immediate traces.
The produced output sequence encodes the predicted traces. 
To generate more future traces, we use the last trace from the training text file concatenated with the two newly generated predicted traces as the next input sequence to the Seq2Seq model to generate the next two future traces.

It is necessary to note that during training, at a state where the DLM weights may be updated to encounter repetition problems, the predicted activities are simply repeating the same activity until the model runs out of the maximum available tokens specified in the training settings without generating the ``$\emph{EOT}$’’ token.
There are also times that the prediction itself gets too long such that it runs out of the maximum available tokens, which also results in the DLM producing incomplete traces. 
In either case, we take the ``$\emph{EOT}$’’ token as the indication of a complete trace, truncate all the incomplete traces, and append the complete traces to the prediction log. 
We store the predicted traces in a text file and remove the ``$\emph{EOT}$’’ tokens at the end of the predicted traces.
The output file is formatted the same way as the input file; it contains multiple lines, where each line records a predicted trace composed of several activities separated by whitespace characters.

\subsection{Prediction Evaluation}
\label{stage5}
In the final stage, a fair, suitable measurement should be applied to assess the quality of the prediction or a post-processed prediction. 
In this work, we use weighted adjacency matrices to compare the ground truth and predicted log.
Specifically, to compare two event logs, we construct their weighted adjacency matrices and then quantify the discrepancies between them.
If the prediction is close to the ground truth, the difference between the two matrices should also be small.
We used MAE and RMSE to measure the differences, as explained in \cref{Preliminaries}. 
The selected measures of the prediction quality are fair to the DLM, as the model was trained and used to predict the next token in the log, and the MAE and RMSE measures quantify the differences in the ground truth and predicted directly-follows relationships.
Note that similar measures were used in other process forecasting works~\cite{Smedt2023}.

There are no widely accepted benchmarks for event log prediction. 
Consequently, we designed three baselines to evaluate our approach: 
(i) every future trace is predicted to be the most frequent historical trace variant (HighestFreq);
(ii) the future traces are predicted randomly based on the uniform distribution of the historical trace variants observed in the training data (RandomPred); and 
(iii) the future traces are predicted with the weighted probability based on the distribution of the historical trace variants observed in the training data (WeightedProb).
To mitigate the impact of a random prediction in the RandomPred and WeightedProb approaches, we made one hundred predictions for each approach to obtain the average and standard deviation for each measurement.

\vspace{-2mm}
\section{Results and Analysis}
\label{eval}
\vspace{-1mm}
This section presents and discusses the results of our log prediction approach.
\subsection{Results}
\label{results}
Our implementation of the PELP approach to event log prediction and the code for replicating the experiments reported in this section are publicly available.\footnote{\url{https://github.com/zhoudayun81/PELP.git}}

To test the effectiveness of the DL model prediction, we first used synthetic event logs that display perfect seasonality of a combination of different trace characteristics, such as parallel processes (for example, $\langle a, b, c, d\rangle$ and $\langle a, c, b, d\rangle$), longer and shorter loops (for instance, $\{\langle a, b, c, d\rangle, \langle a, b, c, b, c, d \rangle\}$ and $\{\langle a, b, d\rangle, \langle a, b, b, d\rangle \}$), and skipped process activities (for example, $\langle a, b, d\rangle$ and $\langle a, d\rangle$).
We also varied the seasonal change from ``short’’ to ``long’’, that is, from two traces $([\langle a, b, c, d\rangle^2,$ $\langle a, c, b, d\rangle^2],$ $\ldots)$ to five traces $([\langle a, b, c, d\rangle^5,$ $\langle a, c, b, d\rangle^5],$ $\ldots)$. 
We also used logs of seasonal changes between three trace variants, such as $([\langle a, b, c\rangle^2,$ $\langle a, b, b, c\rangle^2,$ $\langle a, c\rangle^2],$ $\ldots)$. 
The results over the constructed 18 synthetic event logs demonstrated that given enough training time (epochs), and by using suitable trace concatenation windows, the model can learn the event log pattern in perfection; that is, \emph{it can predict precisely the ground truth future traces in the correct order for all the datasets}. 
It requires more than 200 epochs to reach zero loss for stop training, while for real-life logs, it requires more than 500 epochs to get satisfactory results. We trained our model with the stop condition which only stops after no loss improvement for 100 epochs.

As DLM showed its effectiveness in predicting seasonality in synthetic event logs, we also tested it on seven real-life event logs to evaluate its effectiveness in practice.
\cref{prediction_mae_rmse} summarizes the quality of the logs predicted by the baseline and Seq2Seq DLM approaches. 
The results for the Seq2Seq models are based on the best models selected using the hyper-parameter search described in \cref{stage3}. 
Recall that we made one hundred predictions for RandomPred and WeightedProb baselines and obtained the average and standard deviation measurements.

\begin{table}[t]
    \vspace{-8pt}
\centering
\caption{MAE and RMSE of Seq2Seq DLM event log predictions.}
\label{prediction_mae_rmse}
\smallskip
\begin{tabular}{|l|rrrr|}
\hline
\multirow{2}{*}{Event log} & \multicolumn{4}{c|}{RMSE} \\
\cline{2-5}
& \,\,\,HighestFreq\,\,\, & \,\,\,RandomPred\,\,\, & \,\,\,WeightedProb\,\,\, & \,\,\,\,\,\,PELP\,\,\,\,\,\,\\
\hline
BPIC2011 & 11.65\,\,\,\,\,\, & {\bfseries 3.40±0.43}\,\,\, & 4.19±0.48\,\,\, & 7.80\,\,\,\,\,\, \\
BPIC2012 & 47.52\,\,\,\,\,\, & 38.17±3.93\,\,\, & {\bfseries 9.92±2.98}\,\,\, & {\bfseries 10.20}\,\,\,\,\,\, \\
BPIC2017 & 52.36\,\,\,\,\,\, & 17.81±3.57\,\,\, & {\bfseries 9.10±3.18}\,\,\, & 15.04\,\,\,\,\,\, \\
BPIC2019 & 26.38\,\,\,\,\,\, & 51.43±21.09\,\,\, & {\bfseries 13.18±4.43}\,\,\, & 24.72\,\,\,\,\,\, \\
HB       & 15.69\,\,\,\,\,\, & 13.96±1.32\,\,\, & {\bfseries 1.80±0.54}\,\,\, & {\bfseries 1.78}\,\,\,\,\,\, \\
Sepsis   & 32.01\,\,\,\,\,\, & 12.17±4.72\,\,\, & {\bfseries 5.87±2.67}\,\,\, & {\bfseries 3.77}\,\,\,\,\,\, \\
RTFMP    & 11.17\,\,\,\,\,\, & 15.94±0.83\,\,\, & {\bfseries 2.56±0.82}\,\,\, & 3.88\,\,\,\,\,\, \\
\hline
\multirow{2}{*}{Event log} & \multicolumn{4}{c|}{MAE} \\
\cline{2-5}
& \,\,\,HighestFreq\,\,\, & \,\,\,RandomPred\,\,\, & \,\,\,WeightedProb\,\,\, & \,\,\,\,\,\,PELP\,\,\,\,\,\,\\
\hline
BPIC2011 & 0.30\,\,\,\,\,\, & {\bfseries 0.10±0.01}\,\,\, & 0.11±0.01\,\,\, & 0.21\,\,\,\,\,\, \\
BPIC2012 & 7.76\,\,\,\,\,\, & 5.96±0.53\,\,\, & {\bfseries 1.33±0.26}\,\,\, & {\bfseries 1.39}\,\,\,\,\,\, \\
BPIC2017 & 9.62\,\,\,\,\,\, & 2.70±0.41\,\,\, & {\bfseries 1.60±0.36}\,\,\, & 2.51\,\,\,\,\,\, \\
BPIC2019 & 3.74\,\,\,\,\,\, & 5.52±1.83\,\,\, & {\bfseries 1.85±0.42}\,\,\, & 3.23\,\,\,\,\,\, \\
HB       & 3.78\,\,\,\,\,\, & 3.65±0.27\,\,\, & {\bfseries 0.52±0.13}\,\,\, & {\bfseries 0.51}\,\,\,\,\,\, \\
Sepsis   & 8.31\,\,\,\,\,\, & 2.87±0.68\,\,\, & {\bfseries 1.66±0.39}\,\,\, & {\bfseries 1.40}\,\,\,\,\,\, \\
RTFMP    & 2.88\,\,\,\,\,\, & 5.64±0.27\,\,\, & {\bfseries 0.77±0.21}\,\,\, & 1.22\,\,\,\,\,\, \\
\hline
\end{tabular}
    \vspace{-16pt}
\end{table}
\subsection{Analysis}
\label{Analysis}
In \cref{prediction_mae_rmse}, the best performing RMSE and MAE scores are highlighted with boldface. 
The results are consistent between RMSE and MAE.
The WeightedProb approach achieves good results across the datasets.
Seq2Seq DLMs achieve comparable performance and perform better for some event logs.
Note that the WeightedProb approach reports the average result of multiple predictions, while the Seq2Seq DLM is guaranteed to demonstrate the reported quality level.

\begin{figure}[t]
\vspace{-24pt}
\includegraphics[width=\textwidth]{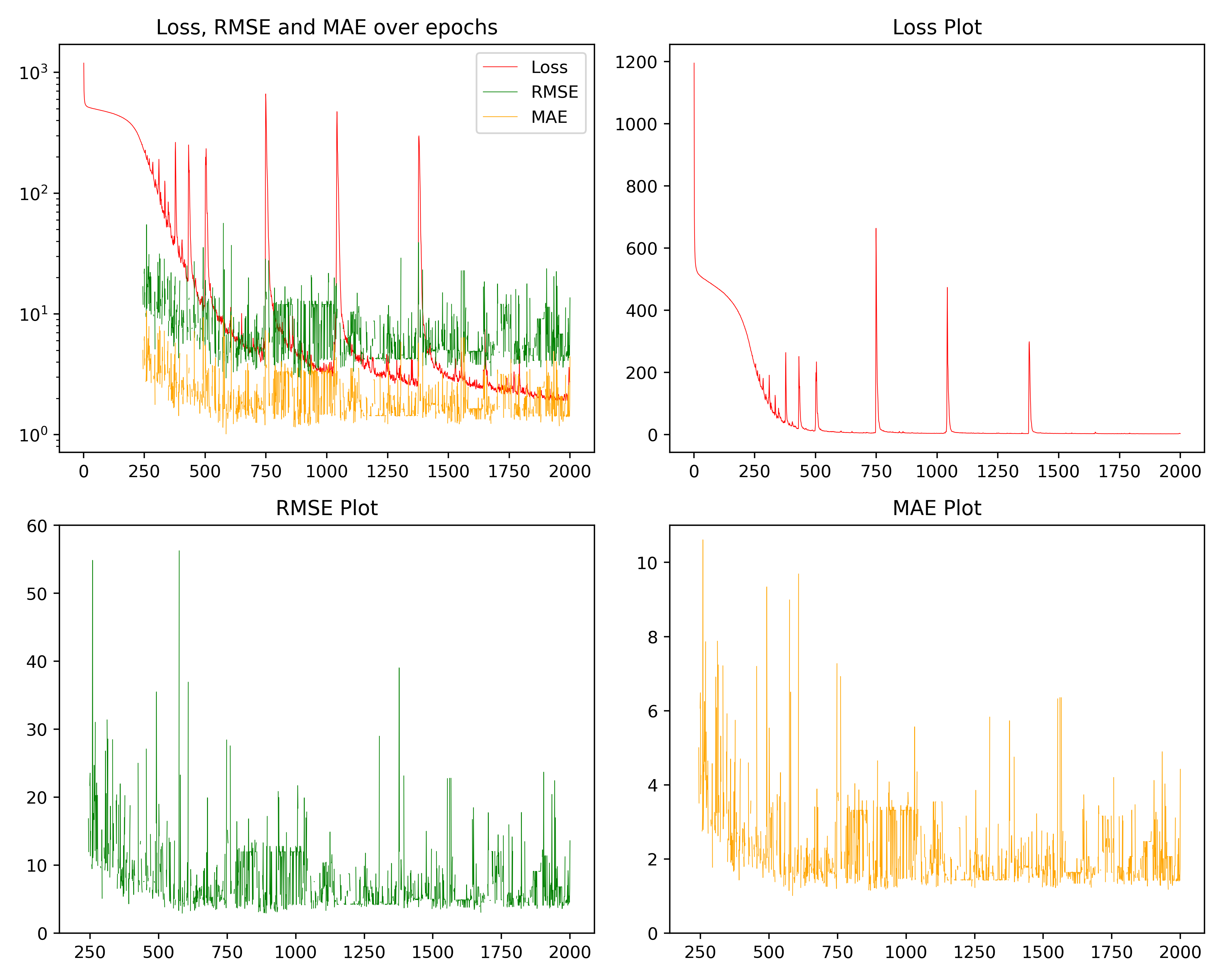}
\caption{RMSE and MAE over epochs on Sepsis log.} 
\label{fig:rmse-mae-over-time}
\vspace{-24pt}
\end{figure}

We observed that Seq2Seq DLMs learn their prediction strategy gradually.
At the early stages of training, the generated predictions coincide with those by the HighestFreq baseline. 
Then, Seq2Seq models start to predict future logs similar to those produced by the WeightedProb baseline. 
For most datasets, the results are collected from models trained for over one thousand epochs. 
While there are concerns that such DLMs may be over-fitting the data, we observed that the over-trained DLMs have a benign over-fitting effect.
For example, \cref{fig:rmse-mae-over-time} reports the performance of one DLM trained on the Sepsis log for two thousand epochs.
Despite several observed spikes in the loss function, the RMSE and MAE results are reliable as they fluctuate within a reasonable range.

We examined event logs where Seq2Seq DLMs performed poorly, namely, BPIC2011, BPIC2017, and BPIC2019. 
Compared to other datasets, BPIC2011 and BPIC2017 have high average trace lengths, 133.59 and 39.93, respectively. 
While the BPIC2019 log has much shorter average trace lengths of 14.83, it contains much more unique activities and trace variants, 213 and 33, respectively. 
A larger average or median trace length hinders the training process by dramatically increasing the training time because the model maximum neurons in the GRU layer are significantly increased, which makes the DLM harder and longer to train. 
We analyzed the training time over the seven training logs for different hyper-parameter combinations and found that the time can vary from 2.2 seconds to 23.43 hours for training one epoch. 
Hence, we conclude that the Seq2Seq DLM architecture described in this paper is less suitable for the event logs with high trace variance or high average trace length. 
In addition, the selected DLM does not have the ability to predict new activities in the future.
Hence, the DLM can only output activities that it learned during training; if there are new activities that emerge outside the data selected for training the DLM, the model cannot provide such a prediction. 
For BPIC2011 and RTFMP logs, there are 60 and 1 activities that only exist in the testing dataset but not in training, each contributes to 0.23 percent and 0.11 percent reduction of accuracy, respectively.

\begin{figure}[t]
\vspace{-24pt}
\centering
\subcaptionbox{\scriptsize Sepsis $\mathit{LacticAcid} \to \mathit{Leucocytes}$.
\label{fig:sepsis-autocorrelation}}{
\includegraphics[width=0.4\linewidth]{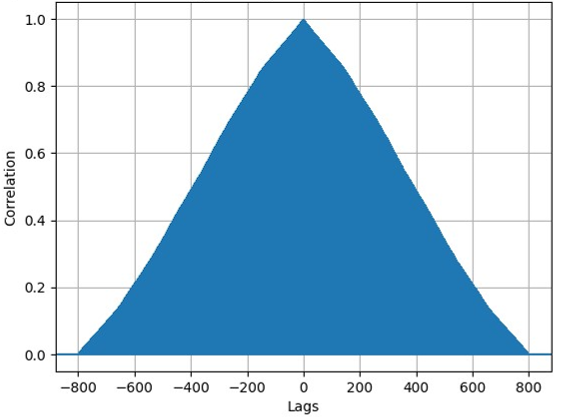}
}
\hspace{34pt}
\subcaptionbox{\scriptsize BPIC2019 $\mathit{CancelGoodsReceipt} \to \mathit{RecordServiceEntrySheet}$.
\label{fig:bpic2019autocorrelation}}{
\includegraphics[width=0.4\linewidth]{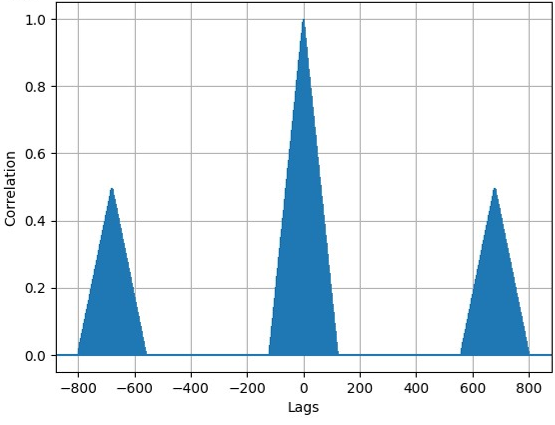}
}
\vspace{-8pt}
\caption{\small{Auto-correlation of directly-follows relationships (first one thousand traces).}}
\label{fig:autocorrelations}
\vspace{-18pt}
\end{figure}

To further explain the results of Seq2Seq DLMs, we analyzed the auto-correlations of the weighted adjacency matrices over time.
For an event log, there are a total of $N \times N$ directly-follows relationships to be analyzed, where $N$ represents the number of distinct activities encountered in that log. 
\cref{fig:sepsis-autocorrelation} demonstrates an example auto-correlation of directly-follows relationship from the \emph{LacticAcid} activity to the \emph{Leucocytes} activity in the Sepsis event log with a window of two hundred cases, and the window moves one case per lag. 
While the majority of the auto-correlations are similar to the one shown in \cref{fig:bpic2019autocorrelation}, BPIC2019 directly-follows relationship from \emph{CancelGoodsReceipt} to \emph{RecordServiceEntrySheet} does not have a high auto-correlation with the beginning of the process collection time. 
\cref{fig:process-model-structure-change-over-time} also shows that the process model at the last time window for prediction has little similarities to the model at the beginning or any interim stage. 
We conclude that for the event logs in which the majority of direct-follows relationships do not have strong auto-correlations over time, our Seq2Seq DLM struggles to provide satisfactory prediction results; otherwise, Seq2Seq DLM or the WeightedProb baseline delivers the best results.

\vspace{-3mm}
\section{Conclusion}
\label{Conclusion}
\vspace{-2mm}
In this paper, we proposed a solution to the event log prediction problem. 
Specifically, we built and trained a Seq2Seq deep learning model to predict event logs. 
The results in synthetic data showed that the Seq2Seq model has the potential to learn seasonality and trend.
Further fine-tuning of the model can improve its performance,  with issues we articulated in \cref{Analysis} aiming to extract the potential of the deep learning approach. 
A solution directly transferable to event log prediction in a general case has yet to come. 
Overall, it is still interesting to investigate deep learning models and different techniques during the five stages of event log prediction for a more accurate prediction, which will help business planning and process change more smoothly.
%
\vspace{-2mm}
\bibliographystyle{splncs04}
\bibliography{bibliography}

\begin{thebibliography}{10}
\providecommand{\url}[1]{\texttt{#1}}
\providecommand{\urlprefix}{URL }
\providecommand{\doi}[1]{https://doi.org/#1}

\bibitem{RN140}
van~der Aalst, W.M.P., Schonenberg, M.H., Song, M.: Time prediction based on process mining. Information Systems  \textbf{36}(2),  450--475 (2011)

\bibitem{DBLP:books/sp/Aalst16}
van~der Aalst, W.M.P.: Process Mining - Data Science in Action, Second Edition. Springer (2016)

\bibitem{bahdanau2016neural}
Bahdanau, D., Cho, K., Bengio, Y.: Neural machine translation by jointly learning to align and translate (2016)

\bibitem{firstActivityPrediction}
Cardoso, J., Lenič, M.: Web process and workflow path mining using the multimethod approach. IJBIDM  \textbf{1}(3),  304--328 (2006)

\bibitem{che2016recurrent}
Che, Z., Purushotham, S., Cho, K., Sontag, D., Liu, Y.: Recurrent neural networks for multivariate time series with missing values (2016)

\bibitem{RN1058}
Di~Francescomarino, C., Ghidini, C., Maggi, F.M., Petrucci, G., Yeshchenko, A.: An eye into the future: Leveraging a-priori knowledge in predictive business process monitoring (2017)

\bibitem{RN50}
Evermann, J., Rehse, J.R., Fettke, P.: Predicting process behaviour using deep learning. Decision Support Systems  \textbf{100},  129--140 (2017)

\bibitem{Hodson:2022aa}
Hodson, T.O.: Root-mean-square error (rmse) or mean absolute error (mae): When to use them or not. Geoscientific Model Development  \textbf{15},  5481--5487 (2022)

\bibitem{10267858}
of~the IEEE Computational Intelligence~Society, S.C.: Ieee standard for extensible event stream (xes) for achieving interoperability in event logs and event streams. IEEE Std 1849-2023 (Revision of IEEE Std 1849-2016) pp. 1--55 (2023)

\bibitem{RN565}
Jalayer, A., Kahani, M., Beheshti, A., Pourmasoumi, A., Motahari-Nezhad, H.R.: Attention mechanism in predictive business process monitoring. In: IEEE 24th EDOC. pp. 181--186 (2020)

\bibitem{RN2484}
Le, M., Gabrys, B., Nauck, D.: A hybrid model for business process event prediction. In: 32nd SGAI. pp. 179--192. Springer London (2012)

\bibitem{RN656}
Metzger, A., Leitner, P., Ivanović, D., Schmieders, E., Franklin, R., Carro, M., Dustdar, S., Pohl, K.: Comparing and combining predictive business process monitoring techniques. IEEE Trans. Syst. Man Cybern. Syst.  \textbf{45}(2),  276--290 (2015)

\bibitem{DBLP:conf/bpm/PollPRRR18}
Poll, R., Polyvyanyy, A., Rosemann, M., R{\"{o}}glinger, M., Rupprecht, L.: Process forecasting: Towards proactive business process management. In: {BPM}. LNCS, vol. 11080, pp. 496--512. Springer (2018)

\bibitem{DBLP:conf/er/SmedtYPWM21}
Smedt, J.D., Yeshchenko, A., Polyvyanyy, A., Weerdt, J.D., Mendling, J.: Process model forecasting using time series analysis of event sequence data. In: {ER}. LNCS, vol. 13011, pp. 47--61. Springer (2021)

\bibitem{Smedt2023}
Smedt, J.D., Yeshchenko, A., Polyvyanyy, A., Weerdt, J.D., Mendling, J.: Process model forecasting and change exploration using time series analysis of event sequence data. Data Knowl. Eng.  \textbf{145},  102145 (may 2023)

\bibitem{DBLP:conf/nips/SutskeverVL14}
Sutskever, I., Vinyals, O., Le, Q.V.: Sequence to sequence learning with neural networks. In: {NIPS}. pp. 3104--3112 (2014)

\bibitem{10.5555/2969033.2969173}
Sutskever, I., Vinyals, O., Le, Q.V.: Sequence to sequence learning with neural networks. In: 27th International Conference on Neural Information Processing Systems - Volume 2. p. 3104–3112. NIPS'14, MIT Press, Cambridge, MA, USA (2014)

\bibitem{DBLP:journals/topnoc/SyringTA19}
Syring, A.F., Tax, N., van~der Aalst, W.M.P.: Evaluating conformance measures in process mining using conformance propositions. Trans. Petri Nets Other Model. Concurr.  \textbf{14},  192--221 (2019)

\bibitem{RN1572}
Tax, N., Verenich, I., La~Rosa, M., Dumas, M.: Predictive business process monitoring with lstm neural networks (2017)

\bibitem{RN1584}
Tschumitschew, K., Nauck, D., Klawonn, F.: A classification algorithm for process sequences based on markov chains and bayesian networks (2010)

\bibitem{RN2863}
Verenich, I.: A general framework for predictive business process monitoring. In: Rinderle-Ma, S., Pastor, O., Wieringa, R. (eds.) CAiSE 2016 Doctoral Consortium. vol.~1603. CEUR-WS (2016)

\bibitem{yu2020hyperparameter}
Yu, T., Zhu, H.: Hyper-parameter optimization: A review of algorithms and applications (2020)

\end{thebibliography}

\end{document}